# AI4PLE: A Methodology for Integrating AI into Product Line Engineering

**Bedir Tekinerdogan**
Information Technology, Wageningen University & Research
Wageningen, The Netherlands
bedir.tekinerdogan@wur.nl

**Abstract**

Reuse-based development has become increasingly important in the creation of complex systems, offering significant opportunities to reduce costs, improve quality, and accelerate time-to-market. Product Line Engineering (PLE) provides a systematic approach to realizing this potential by enabling the efficient creation, management, and customization of product families by reusing shared assets and capabilities. PLE involves addressing numerous complex decisions, including feature selection, variability management, and configuration optimization, which are critical to the success of a product line. Despite its promise, the systematic integration of Artificial Intelligence (AI) into PLE processes has not yet been comprehensively explored. In this paper, we propose a methodological framework to support the systematic integration of AI into PLE and evaluate its effectiveness through a multi-case study conducted in an industrial context.

## 1. Introduction

Modern engineering practices are increasingly recognizing the vast potential for systematic reuse in the development of complex systems. Many systems share significant commonalities, with only a small subset of their features or components being unique to specific products. This observation underscores the opportunity to leverage reuse at various levels, ranging from individual components and modules to large-scale reuse of architectures, processes, and assets across entire product families. Large-scale reuse is particularly impactful, as it focuses on creating shared platforms and managing variability to address the needs of multiple products efficiently. A reuse-based development approach aims to minimize redundancy and enables organizations to exploit and benefit from the commonalities across systems while tailoring solutions for diverse requirements.

Product Line Engineering (PLE) has emerged as a systematic methodology for achieving large-scale reuse in both software and systems engineering [1][7]. By enabling the creation, management, and evolution of product families through the reuse of shared assets and the management of variability, PLE supports organizations in streamlining development processes and reducing inefficiencies. The benefits of adopting PLE are well-documented, with organizations reporting significant improvements in productivity, reduced time-to-market, enhanced product quality, increased customer satisfaction, and better utilization of resources. Furthermore, PLE supports mass customization, enabling organizations to respond to market demands while maintaining a competitive edge in dynamic environments.

Central to PLE's success is its decision-making process, which spans activities such as feature selection, variability modeling, and configuration optimization. These decisions are critical to ensuring the efficiency, scalability, and adaptability of product lines. However, as product families become complex, the challenges of managing variability, interdependencies, and scalability increase exponentially. Addressing these challenges requires advanced methods to support decision-making while maintaining efficiency and consistency across the PLE lifecycle.

Artificial Intelligence (AI) has shown great promise in addressing such challenges. Techniques like machine learning, optimization algorithms, and natural language processing have demonstrated their ability to automate complex tasks, analyze large datasets, and improve decision-making processes in engineering contexts. Within PLE, AI offers significant potential to enhance critical lifecycle activities by improving decision-making and supporting the human engineer. AI also supports the automation of labor-intensive tasks, allowing teams to focus on strategic activities while maintaining high levels of efficiency. These capabilities position AI as a transformative tool for advancing PLE practices and achieving greater operational effectiveness. Despite its potential, the systematic integration of AI into PLE has not been comprehensively explored. Current approaches often lack the structured methodologies needed to align AI techniques with the specific demands of PLE effectively. This gap presents an opportunity to develop frameworks that bridge these domains, enabling organizations to harness the strengths of AI to optimize product line engineering processes.

The remainder of the paper is organized as follows. Section II outlines the research methodology used to develop the AI4PLE framework. Section III discusses the foundational concepts of PLE and its two-lifecycle process. Section IV elaborates on the key components of the AI4PLE framework. Section V presents a multi-case study validating the proposed approach. Finally, Section VI summarizes the contributions and implications of this research.

## 2. Research Methodology

Our methodology for developing and applying the AI-SoS framework adopts a structured approach, ensuring that AI techniques are effectively integrated into the lifecycle and challenges of PLE. This section delineates the methodology for the development of the AI4PLE framework. Fig 1. shows the adopted method for developing the AI4PLE framework. The development of the AI4PLE framework follows a structured methodology, focusing on integrating AI techniques into specific steps of the PLE lifecycle:

- *Domain Analysis of AI Technology Space*

This step examines the PLE lifecycle, particularly decision-intensive activities like feature modeling, variability management, and product configuration. Challenges like scalability, consistency, and trade-offs are identified, revealing opportunities for AI to enhance efficiency and decision-making.

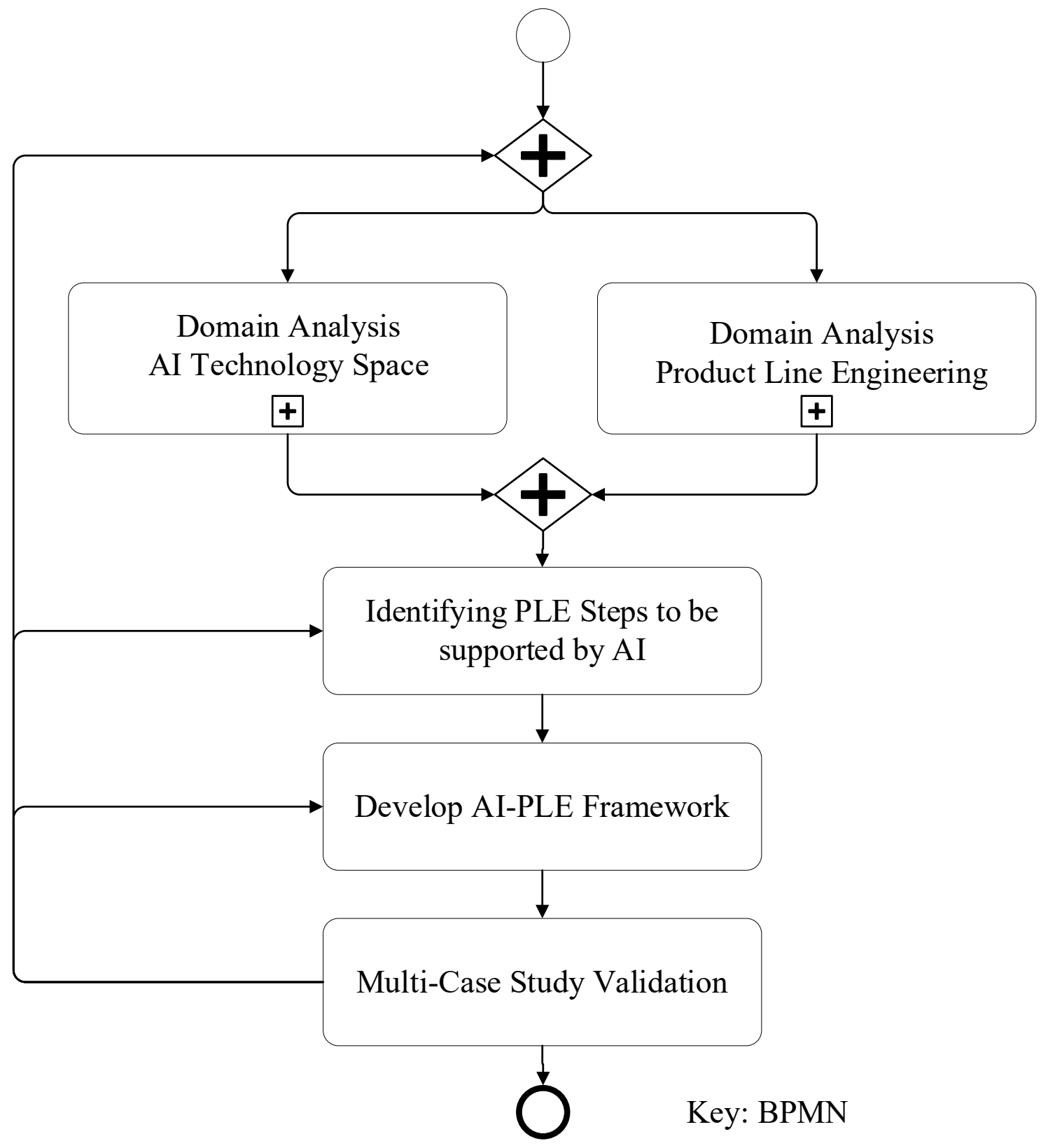

**Fig 1**. Adopted research method for the development of the AI4PLE Framework

- *Analysis of PLE Lifecycle Steps*

The PLE lifecycle is analyzed in detail, focusing on processes such as domain scoping, requirements engineering, design, implementation, testing, and configuration management. This analysis identifies lifecycle steps that involve significant complexity, uncertainty, or repetitive tasks, ensuring AI is targeted to areas of greatest impact.

- *Identifying Lifecycle Steps for AI Integration*

Key lifecycle steps where AI can provide the most value are identified, such as managing variability, optimizing feature selections, and automating testing. AI is applied to enhance decision-making, streamline repetitive processes, and address scalability challenges across the lifecycle.

- *Development of the AI4PLE Framework*

Using insights from the previous steps, the framework is developed to guide AI integration into PLE processes. It includes a metamodel to link AI techniques to lifecycle steps, a matrix chart for mapping AI approaches to specific tasks, and selection guidelines to help practitioners adopt the appropriate AI strategies.

- *Multi-Case Study Validation*

Case studies across industries, such as automotive and software-intensive systems, validate the framework's effectiveness. These studies demonstrate improvements in decision quality, scalability, and overall efficiency when AI is integrated into PLE lifecycle steps.

- *Iterative Refinement*

The framework is refined iteratively to adapt to advancements in AI and changes in PLE practices. Feedback from case studies ensures continuous improvement, maintaining the framework's relevance and usability.

## 3. Product line engineering

Compared to single-system development, adopting a PLE approach requires an upfront investment in establishing reusable platforms, feature models, and supporting processes. However, this investment typically pays off after developing more than one product, as the cumulative benefits increase with each additional product. The point at which these investments are recovered is called the break-even point. Beyond this point, PLE enables organizations to achieve a significant return on investment (ROI), particularly in terms of scalability and efficiency compared to developing each product independently.

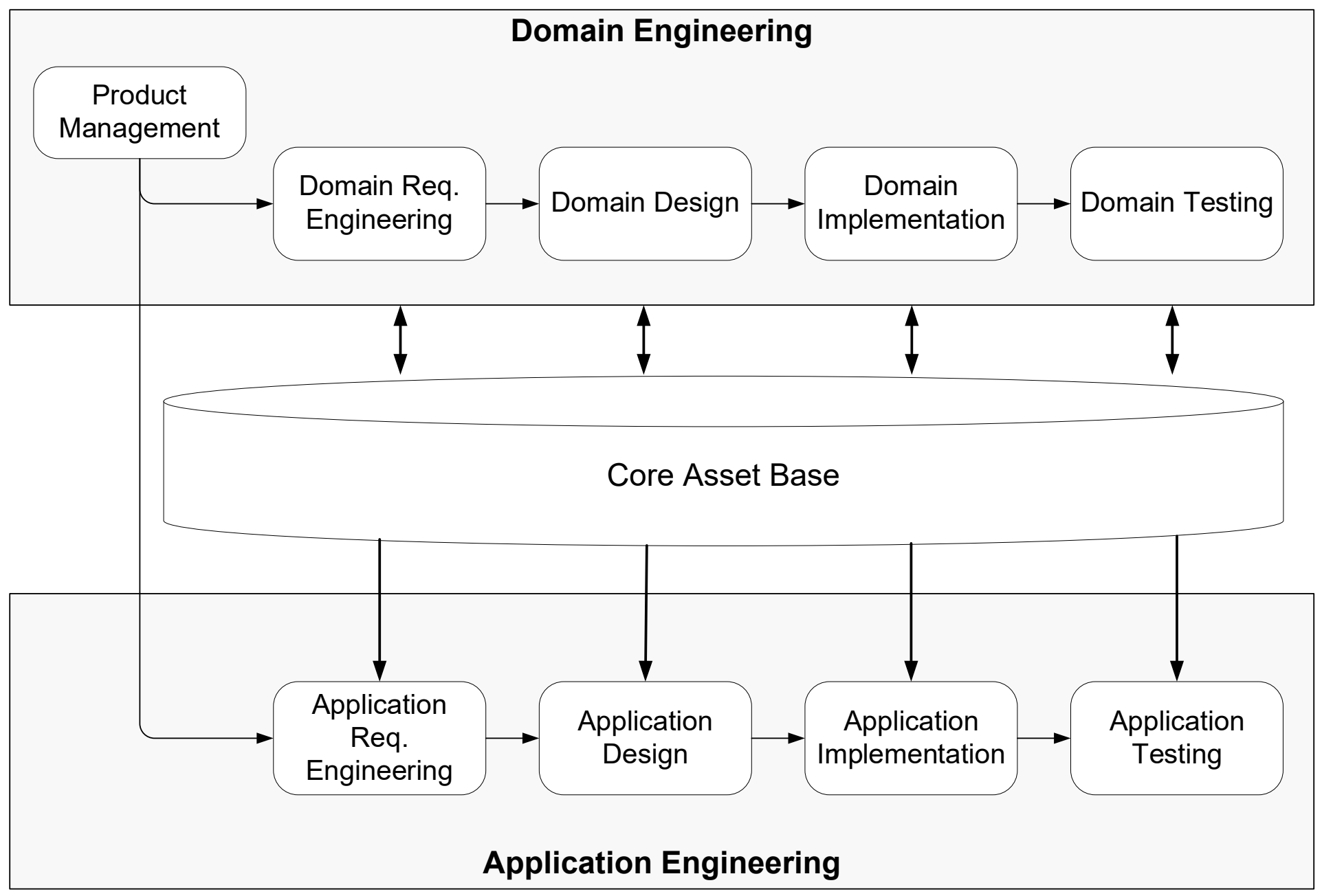

**Fig 2**. Product Line Engineering Lifecycle Process

As shown in Fig 2, the PLE process is structured around two complementary lifecycle activities: domain engineering and application engineering. In domain engineering, the focus is on creating reusable platforms, product line architectures, and variability models that define the commonalities and variabilities across the product family. These assets are then used in application engineering, where individual products are developed by selecting and configuring features based on specific requirements. These lifecycles are supported by a management process that addresses technical and organizational aspects, ensuring alignment with business goals and the effective execution of PLE practices.

PLE is used extensively in software engineering, where shared codebases, libraries, and architectures form the foundation for reusable platforms. However, its principles are equally applicable to systems engineering, where reusable hardware components, interfaces, and subsystems play a similar role. This dual applicability makes PLE a powerful approach for managing variability and achieving reuse across diverse domains, from software-intensive systems to integrated hardware-software solutions.

# 4. AI4PLE Framework

In this section, we present the methodological framework, AI4PLE, for integrating AI into the PLE lifecycle. The framework encompasses several key components: a metamodel representing the relationship between AI techniques and PLE lifecycle activities, the rationale for selecting AI approaches, the criteria for adopting AI techniques, and a structured method for implementing AI in the PLE process. Each of these components will be detailed in the subsequent subsections.

## 4.1 Metamodel

The metamodel that illustrates the concepts for this mapping is displayed in Fig 3. A product line is defined as a set of products developed using a two-phase lifecycle process: domain engineering and application engineering. This PLE lifecycle can be supported by an AI Technology Space, which encompasses AI approaches and AI agents. When adopting AI, various rationales and criteria can be identified to guide its implementation. Lastly, the integration of AI can involve different roles for humans, depending on the level of interaction: in-the-loop, on-the-loop, or out-of-the-loop.

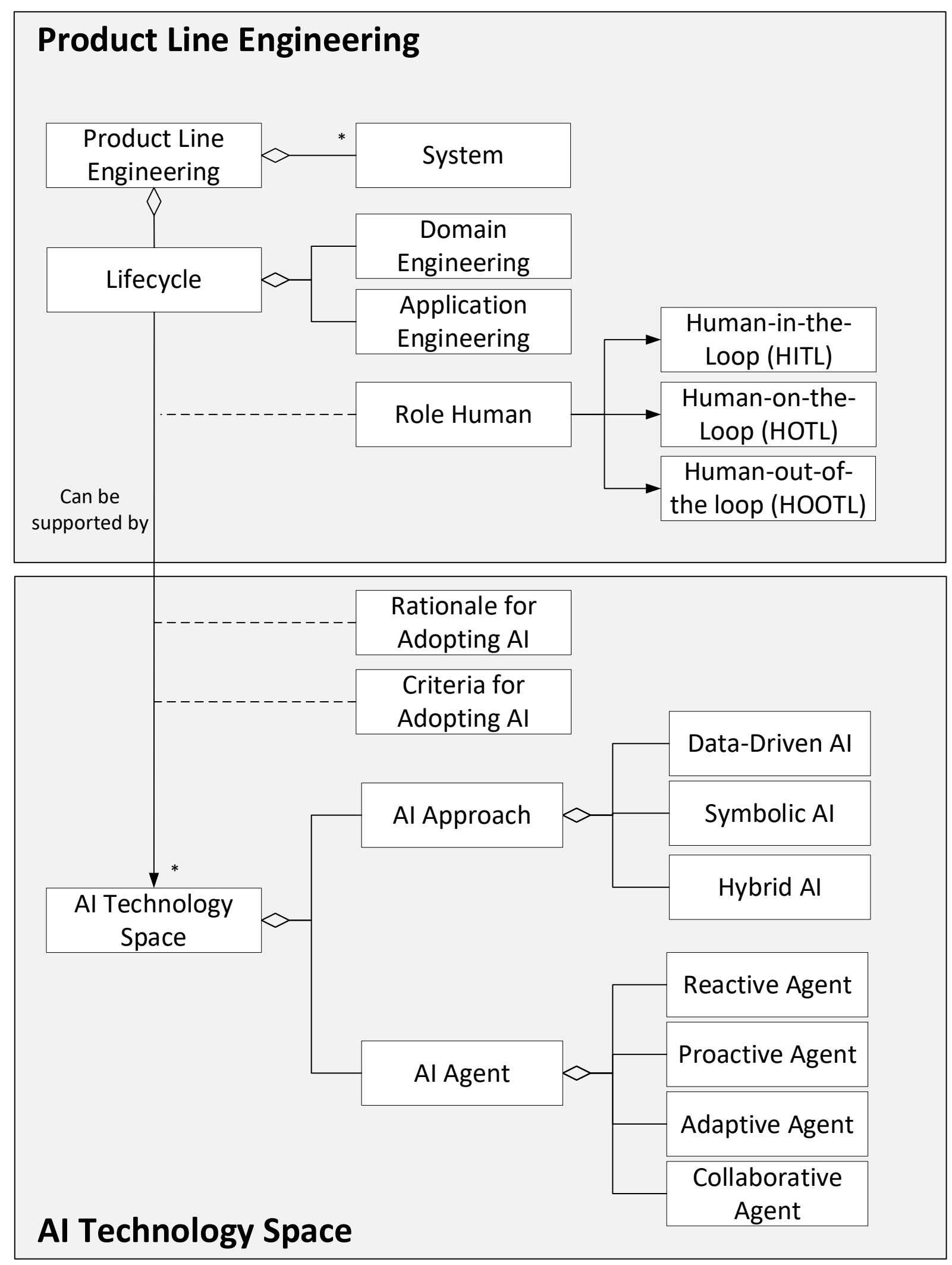

**Fig 3**. Metamodel for AI for Product Line Engineering

## 4.2 Rationale for Selection of AI Approaches in PLE

The selection of AI approaches for integration into the PLE lifecycle is driven by their capacity to address key challenges and opportunities within the process. Each approach is tailored to specific needs, ensuring the technology enhances efficiency, decision-making, and scalability across the lifecycle. The rationale behind these choices can be elaborated as follows:

- *Automating Repetitive and Time-Consuming Tasks*

PLE involves numerous repetitive tasks, such as analyzing requirements, validating consistency across models, and generating test cases. When conducted manually, these tasks are labor-intensive and prone to human error. AI approaches excel at automating such processes, enabling faster and more accurate execution. By automating routine activities, AI reduces the cognitive and operational burden on engineers, allowing them to focus on strategic decisions that enhance product line outcomes. This automation significantly accelerates the PLE workflow while ensuring greater precision and reliability.

- *Enhancing Decision-Making*

Decision-making is a critical aspect of PLE, where engineers must address complex trade-offs between cost, quality, and performance while navigating a highly variable landscape. AI technologies, such as optimization algorithms and predictive analytics, support these decisions by providing actionable insights and simulating potential outcomes. These tools enable stakeholders to make informed, data-driven decisions that effectively balance competing priorities. By leveraging AI, decision-making within PLE becomes more transparent, precise, and scalable, ensuring alignment with organizational goals and market demands.

- *Managing Complexity*

As product lines expand and diversify, managing variability and interdependencies grows exponentially more complex. PLE must address this complexity while ensuring coherence across products and configurations. AI approaches such as clustering, knowledge graphs, and machine learning models are particularly suited to managing large-scale interdependencies and variability. These tools help organizations understand relationships between features, detect inconsistencies, and adapt strategies dynamically as requirements evolve. AI-driven complexity management ensures that the increasing scale of PLE does not compromise efficiency or quality.

- *Improving Efficiency and Scalability*

PLE requires organizations to manage large-scale product families while maintaining flexibility and responsiveness to market changes. Traditional methods often struggle to keep up with the scalability demands of modern product lines. AI addresses these limitations by automating key processes, optimizing resource allocation, and providing real-time insights into performance and variability. The scalability enabled by AI ensures that organizations can handle diverse product families effectively, adapt quickly to changing market conditions, and maintain a competitive edge.

- *Supporting Consistency and Traceability*

Ensuring consistency and traceability across features, configurations, and requirements is a core challenge in PLE. AI-powered tools, such as dependency analysis algorithms and knowledge graphs, can automatically validate relationships and maintain alignment across the product line. These tools enhance traceability by linking requirements to features and configurations, ensuring that changes or updates are consistently reflected throughout the lifecycle. AI reduces the risk of inconsistencies and simplifies the process of managing complex interdependencies.

- *Enabling Proactive Quality Assurance*

Quality assurance in PLE often involves anticipating and mitigating potential risks and failures before they manifest. AI techniques, such as defect prediction models and automated testing frameworks, enable proactive quality management. These tools can identify high-risk components, generate comprehensive test cases, and adapt to evolving configurations, ensuring thorough validation of reusable assets and product variants. AI shifts the focus from reactive to proactive quality assurance, reducing risks and enhancing product reliability.

- *Supporting Lifecycle Evolution and Adaptability*

The PLE lifecycle is dynamic, requiring continuous updates to domain assets and derived products to align with changing market conditions, technologies, and customer needs. AI techniques such as reinforcement learning and predictive analytics are selected for their ability to adapt to evolving requirements. During domain engineering, AI can assist in updating reusable assets, while in application engineering, it can recommend changes to configurations based on customer feedback or market trends. This adaptability ensures that the PLE process remains relevant and responsive throughout its lifecycle.

While this list captures the key motivations for selecting AI approaches, it is not exhaustive. Identifying the rationale for AI integration is an integral part of the process, evolving alongside organizational priorities, technological advancements, and domain-specific challenges. This flexibility allows for the inclusion of additional considerations as new opportunities or requirements emerge. The motivations outlined here provide a foundation, ensuring a structured starting point for AI implementation. However, maintaining adaptability ensures that the integration of AI remains responsive and aligned with the dynamic needs of the PLE lifecycle and the broader organizational context.

## 4.3 Criteria for Adopting AI Techniques in PLE

When integrating AI into PLE to enhance specific lifecycle activities, several key criteria must be carefully evaluated to ensure the implementation is effective, value-adding, and aligned with organizational goals. The criteria for adopting AI in PLE are outlined below:

- *Data Availability and Quality*

AI systems rely heavily on the availability of sufficient, high-quality data to generate insights and support decisions. In the context of PLE, this includes data related to requirements, feature models, configurations, and product performance. Ensuring that the data is abundant, accurate, and representative of the variability within the product line is essential for training robust AI models and delivering reliable outcomes. Poor data quality or insufficient data can significantly limit the effectiveness of AI integration.

- *Complexity of the Concern*

PLE often involves managing complex interdependencies among features, variability constraints, and large-scale product configurations. AI is particularly valuable for addressing these multifaceted challenges due to its ability to process large volumes of interconnected data, model intricate relationships, and adapt to dynamic environments. Evaluating whether the concern involves sufficient complexity to benefit from AI-driven analysis and decision-making is a critical criterion for adoption.

- *Repetitiveness and Scalability of Tasks*

Many PLE activities, such as feature extraction, consistency checking, and configuration validation, are repetitive and time-consuming. AI excels in automating these tasks, improving efficiency and reducing human error. Additionally, AI's ability to scale operations is crucial in PLE environments where the number of product variants and configurations grows over time. Tasks that are repetitive and expected to expand in scope are prime candidates for AI integration.

- *Potential for Enhanced Decision-Making through AI*

PLE requires numerous high-stakes decisions, such as selecting features, optimizing trade-offs, and managing variability. AI can significantly enhance these decisions by providing predictive insights, real-time data analysis, and simulations of potential outcomes. When the ability to improve decision-making processes is evident, AI adoption becomes a strategic advantage for organizations managing complex product lines.

- *Regulatory and Compliance Considerations*

In regulated industries, maintaining compliance with legal and industry standards is critical throughout the PLE lifecycle. AI can assist in monitoring compliance by ensuring that configurations, feature selections, and derived products adhere to relevant regulations. This includes automating checks for consistency with standards and maintaining detailed records for auditability. AI's ability to seamlessly integrate with existing systems and processes should be evaluated to ensure alignment with compliance requirements

- *Integration Feasibility with Existing Processes*

AI solutions must align with the existing PLE architecture and processes without causing significant disruptions. Seamless integration ensures that AI tools enhance current workflows rather than complicating or replacing them. Organizations must assess the compatibility of AI technologies with their existing tools for feature modeling, configuration management, and testing, ensuring a smooth adoption process.

- *Cost-Benefit Analysis*

A thorough cost-benefit analysis should guide the decision to adopt AI. This involves evaluating whether the benefits of AI integration—such as improved efficiency, reduced time to market, enhanced decision-making, and scalability—justify the costs associated with its deployment, training, and maintenance. The analysis should also consider long-term gains, such as increased product quality and market agility, compared to the initial investment.

## 4.4 Method for Integrating AI in the PLE Process

Once the metamodel, rationale, and criteria for AI adoption in PLE are established, the next step is to define the method for integrating AI into the PLE lifecycle. Fig 4 shows the generic workflow with the steps for identifying and integrating AI in the PLE lifecycle process.

- *Analyze the PLE Lifecycle*

Identify the steps in the PLE lifecycle process or activities that could most benefit from AI. Prioritize tasks with high complexity, repetitive operations, or scalability challenges.

- *Describe the Rationale for AI Integration*

For each selected step, document the rationale for using AI. Highlight the challenges AI must address, such as improving decision-making, automating routine tasks, or managing variability. Ensure the rationale aligns with organizational goals.

- *Define Criteria for AI Integration*

Establish criteria for integrating AI, such as data availability and quality, automation potential, scalability needs, and technical feasibility. Consider compatibility with existing PLE tools and workflows to ensure smooth implementation.

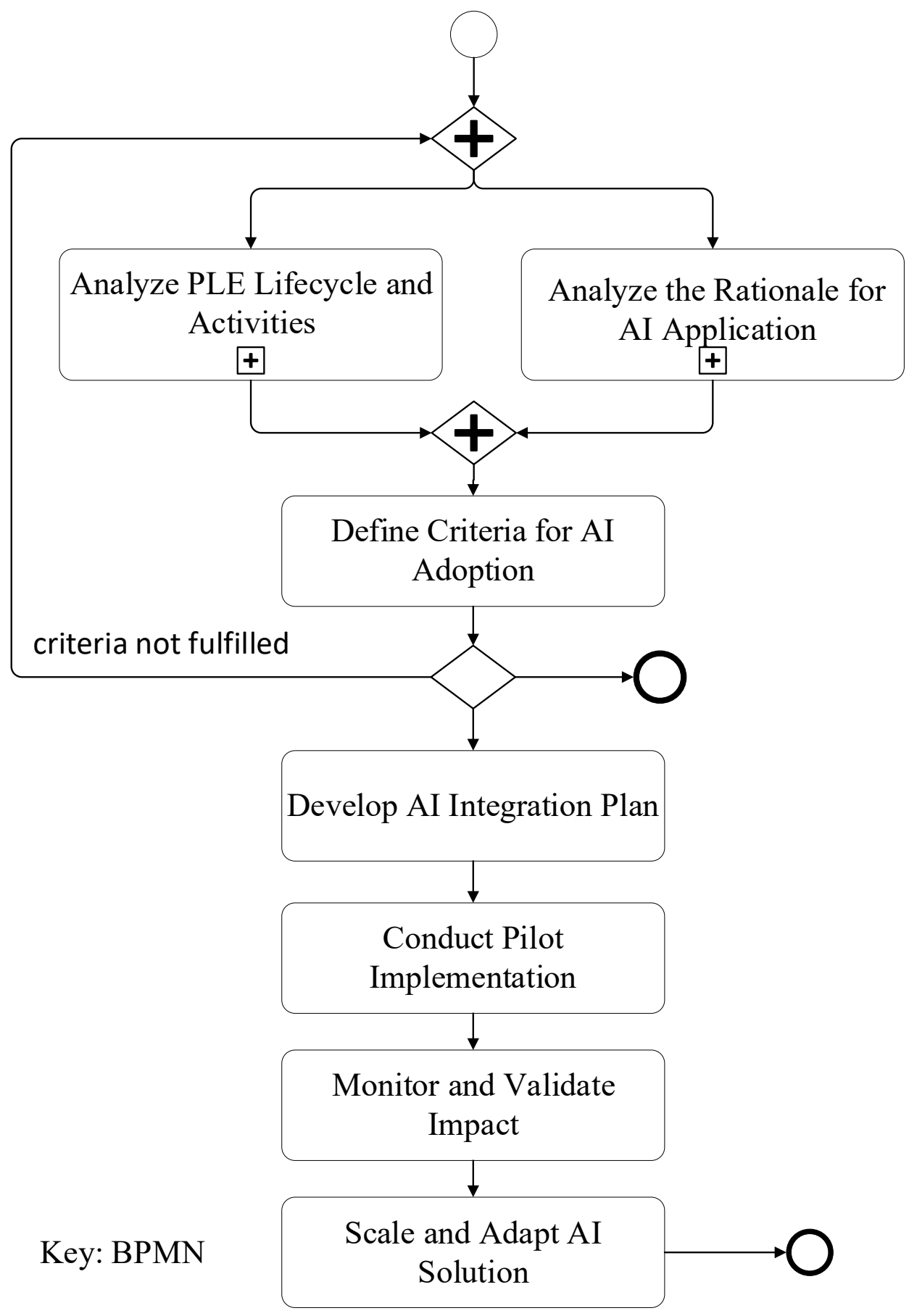

**Fig 4**. Method for integrating AI in PLE

- *Develop an AI Integration Plan*

Create a detailed plan outlining how AI will be incorporated into the selected lifecycle processes. Specify the AI techniques to be applied, such as optimization algorithms or Natural Language Processing, and define performance metrics to evaluate success.

- *Conduct Pilot Implementation*

Apply AI to a specific lifecycle activity in a controlled environment to test feasibility and measure impact. Use this phase to validate assumptions, refine techniques, and address potential challenges before full-scale adoption.

- *Monitor and Validate Impact*

Continuously track the performance of AI integration using metrics such as efficiency, decision accuracy, and scalability. Gather feedback from stakeholders to ensure AI is delivering the expected value.

- *Scale and Adapt AI Solutions*

Extend AI integration to encompass more lifecycle processes and larger product families. Adapt AI systems to handle increased complexity, larger datasets, and evolving organizational needs, ensuring scalability.

- *Iterate and Refine*

Continuously refine the integration process by incorporating feedback, responding to technological advancements, and adapting to changes in PLE practices. Ensure AI remains effective and aligned with organizational objectives.

These steps can be applied to different projects. Although each project will differ in scope and outcomes, the following provides overall guidance for ensuring effective integration.

# 5. Case Study Research

The AI4PLE framework has been applied within a real industrial context on a product line for Physical Protection Systems (PSS) [2][6], which are integrated solutions designed to protect people, assets, and facilities from threats such as unauthorized access, theft, vandalism, and other physical risks. PSS typically include various components, such as access control systems (e.g., card readers, biometric scanners), surveillance equipment (e.g., CCTV cameras), perimeter protection mechanisms (e.g., fences, motion sensors), and intrusion detection systems. By combining hardware, software, and procedural safeguards, PSS ensures the safety and operational continuity of critical infrastructure and resources. The project spanned three years and involved the author's direct participation as a researcher, consultant, and system architect.

To validate the AI4PLE framework, we conducted an empirical case study within an industrial project based on Runeson and Höst's guidelines [4]. This included five key steps: designing the case study, preparing for data collection, executing the study with data collection, analyzing the collected data, and reporting the findings. The primary goal was to assess the impact of integrating AI into the PLE process within an industrial context to evaluate the framework's effectiveness and practicality. A comprehensive stakeholder and domain analysis of the application domain was conducted before the evaluation. This initial phase provided detailed insights into the organizational requirements and challenges specific to the case study, ensuring the framework was applied in a realistic and relevant context.

We applied the AIPLE framework by following the workflow outlined in Fig 4. This involved identifying and implementing AI techniques in domain engineering and application engineering processes. The workflow facilitated a systematic evaluation of how AI could enhance decision-making, manage variability, and automate complex tasks within the product line lifecycle.

From the case study, we determined that the rationale for applying AI to the PSS product line stems from the complexity and variability inherent in managing such systems. PSS involves various components, such as access control, surveillance, perimeter protection, and intrusion detection, each requiring precise coordination and customization for specific use cases. AI can automate and optimize decision-making processes, such as feature selection, variability management, and configuration validation, reducing manual effort and improving accuracy.

Additionally, AI can analyze large-scale data generated by PSS to identify patterns, predict risks, and provide actionable insights, enhancing the system's overall adaptability and resilience. By leveraging AI-driven techniques like machine learning, optimization algorithms, and natural language processing, organizations can streamline product line processes, improve efficiency, and ensure scalability as the product line evolves. Furthermore, AI supports consistency and reusability across shared assets, ensuring higher product quality and reliability. These capabilities make AI a critical enabler for addressing the challenges of PSS PLE while achieving better alignment with dynamic operational and market requirements.

The adoption of AI in the PSS PLE project was evaluated based on the earlier identified criteria, ensuring its alignment with the unique demands of the domain and its potential to deliver meaningful outcomes. *Data availability and quality* were critical enablers, as the extensive data from access control logs, surveillance feeds, and sensor outputs provided a robust foundation for AI-driven insights. The data's abundance and representativeness supported advanced pattern recognition and predictive capabilities. The *complexity of the concerns* within PSS PLE, including managing interdependencies and variability constraints, underscored the value of AI in modeling intricate relationships and addressing dynamic configurations. The project had to deal with *repetitive and scalable tasks*, such as configuration validation and dependency management. The *potential to enhance decision-making* was evident, with AI providing real-time analytics and simulations to optimize trade-offs and variability management. Furthermore, already in earlier projects, AI demonstrated its utility in *regulatory compliance,* automating checks to ensure configurations adhered to relevant standards and maintaining detailed audit trails. The *integration feasibility* of AI was validated through its compatibility with existing tools and workflows, enabling seamless adoption without disrupting current processes. Finally, a *cost-benefit analysis* confirmed the strategic value of AI, with significant improvements in efficiency, decision quality, and scalability outweighing the initial investment and maintenance costs. These criteria collectively demonstrated that AI integration in the PSS PLE process was feasible and transformative, addressing key challenges while enhancing the overall lifecycle management.

After establishing the rationale and criteria for adopting AI, specific potential AI approaches were identified and are summarized in Table 1. While not all these approaches were implemented during the project, they were carefully analyzed and incorporated into the strategic roadmap for future development and integration efforts. Due to space limitations, we cannot explore each application of AI to the PLE steps. However, in the implementation of the AI4PLE framework for PSS PLE, several AI techniques were indeed employed to enhance various stages of the product line engineering process.

Natural Language Processing (NLP) was used during domain scoping and requirements engineering to extract and analyze information from unstructured texts, such as stakeholder inputs and specifications. Machine Learning (ML) algorithms were applied in domain design and testing phases to predict defect-prone areas, analyze dependencies, and optimize variability models based on historical data, thereby improving decision-making processes. Generative AI models automated the creation of reusable assets, such as code snippets and test cases, during the implementation phase, reducing manual effort and increasing consistency. Additionally, automated testing frameworks powered by AI were employed to generate and execute adaptive test cases, ensuring robust quality assurance for both reusable components and product-specific configurations.

**TABLE 1.** AI Approaches mapped to PLE Activities

| **Lifecycle Activity** | **AI Approach** |
|---|---|
| Product Management | AI-driven analytics for monitoring performance, forecasting, and strategic decision support. |
| Domain Scoping | NLP for extracting variability, ML for scope prediction, and clustering for grouping related features. |
| Domain Requirements Engineering | NLP for extracting requirements, knowledge graphs for traceability, and predictive models for validation. |
| Domain Design | ML for defect prediction, optimization for architectural trade-offs, and constraint-solving for managing dependencies. |
| Domain Implementation | Generative AI for reusable components, automation tools for development, and recommendations for design patterns. |
| Domain Testing | Automated testing frameworks and ML models for defect prediction and test prioritization. |
| Configuration Management | AI for version control, dependency analysis, and predictive configuration planning. |
| Application Req. Eng. | NLP for linking requirements, predictive analytics for validation, and knowledge graphs for impact analysis. |
| Application Implementation | Automation tools for product-specific configurations, ensuring consistency and reducing manual effort. |
| Application Design | AI optimization for tailoring architectures and ML for refining configurations. |
| Application Testing | Automated testing for product validation, ML-based defect prediction, and regression testing. |

## 6. Conclusion

This paper presented a structured approach to integrating AI into the lifecycle of PLE by analyzing the PLE lifecycle and identifying key steps where AI can add value. The proposed AI4PLE framework was developed systematically, incorporating insights from domain analysis, lifecycle evaluation, and case studies. Its components, including the metamodel, the rationale, the criteria, and the method, are aligned with lifecycle-specific needs, enabling organizations to automate repetitive tasks, optimize trade-offs, and manage variability effectively. An empirical case study research validation demonstrated the framework's applicability, highlighting its potential to improve scalability, decision quality, and operational efficiency. The application of AI in the project resulted in significant improvements, including enhanced efficiency and scalability by reducing manual effort in the applied lifecycle steps. On the other hand, the research and project results also helped to identify key research domains for integrating AI in PLE. Future research will include implementing the identified AI approaches and applying them to different case studies

## Declarations

### Funding Statement

The author did not receive support from any organization for the submitted work.

### Ethics, Consent to Participate, and Consent to Publish declarations

Not applicable

### Clinical trial number:

not applicable